\documentclass[11pt,a4paper]{article}
\usepackage[hyperref]{acl2017}
\usepackage{times}
\usepackage{latexsym}
\usepackage[pdftex]{graphicx}  
\usepackage{amssymb}
\usepackage{url}

\aclfinalcopy % Uncomment this line for the all SemEval submissions
\title{BB\_twtr at SemEval-2017 Task 4: Twitter Sentiment Analysis with CNNs and LSTMs}

\author{Mathieu Cliche\\
	Bloomberg \\
  {\tt mcliche@bloomberg.net}}

\date{}

\begin{document}

\maketitle

\begin{abstract}
In this paper we describe our attempt at producing a state-of-the-art Twitter sentiment classifier using Convolutional Neural Networks (CNNs) and Long Short Term Memory (LSTMs) networks.  Our system leverages a large amount of unlabeled data to pre-train word embeddings.  We then use a subset of the unlabeled data to fine tune the embeddings using distant supervision.  The final CNNs and LSTMs are trained on the SemEval-2017 Twitter dataset where the embeddings are fined tuned again.  To boost performances we ensemble several CNNs and LSTMs together. Our approach achieved first rank on all of the five English subtasks amongst 40 teams.
\end{abstract}

\section{Introduction}
\label{sec:intro}

Determining the sentiment polarity of tweets has become a landmark homework exercise in natural language processing (NLP) and data science classes.  This is perhaps because the task is easy to understand and it is also easy to get good results with very simple methods (e.g. positive - negative words counting).  The practical applications of this task are wide, from monitoring popular events (e.g. Presidential debates, Oscars, etc.) to extracting trading signals by monitoring tweets about public companies.  These applications often benefit greatly from the best possible accuracy, which is why the SemEval-2017 Twitter competition promotes research in this area. The competition is divided into five subtasks which involve standard classification, ordinal classification and distributional estimation.  For a more detailed description see \cite{SemEval:2017task4}. \\

In the last few years, deep learning techniques have significantly out-performed traditional methods in several NLP tasks \cite{chen2014fast,bahdanau2014neural}, and sentiment analysis is no exception to this trend \cite{rojas2016deep}.  In fact, previous iterations of the SemEval Twitter sentiment analysis competition have already established their power over other approaches \cite{SemEval:2016task4,severyn-moschitti:2015SemEval,deriu-EtAl:2016SemEval}.  Two of the most popular deep learning techniques for sentiment analysis are CNNs and LSTMs.  Consequently, in an effort to build a state-of-the-art Twitter sentiment classifier, we explore both models and build a system which combines both.  \\
% CNNs have been shown to be very successful at several natural language processing tasks (***).  The task they shine the most at is arguably sentence classification (***), which means CNNs should be a great fit for this competition.
%collobert2011natural,socher2012deep,

This paper is organized as follows.  In sec. \ref{sec:desc} we describe the architecture of the CNN and the LSTM used in our system.  In sec. \ref{sec:train} we expand on the three training phases used in our system.  In sec. \ref{sec:trick} we discuss the various tricks that were used to fine tune the system for each individual subtasks.  Finally in sec. \ref{sec:res} we present the performance of the system and in sec. \ref{sec:conc} we outline our main conclusions.

\section{System description}
\label{sec:desc}

%Deep learning approaches to text classification is usually tackled either by a CNN or a recurrent neural network (RNN), for example the long short term memory network (LSTM).  However historically CNNs have usually outperformed RNNs in the Twitter sentiment analysis portion of the SemEval competition.  In our own experiments we also found that while RNNs gave good performances, CNNs were slightly better.  While the internal memory of a LSTM  might make it more suitable for long texts classification, we beleive that the architecture of the CNN is more suited for short texts.  Indeed, the CNN's structure essentially extracts the most-important n-gram in the embedding space, which is why we beleive those systems are good at sentence classification.  \\

\subsection{CNN}

\begin{figure*}[t]
\centering
\includegraphics[width=139.5mm]{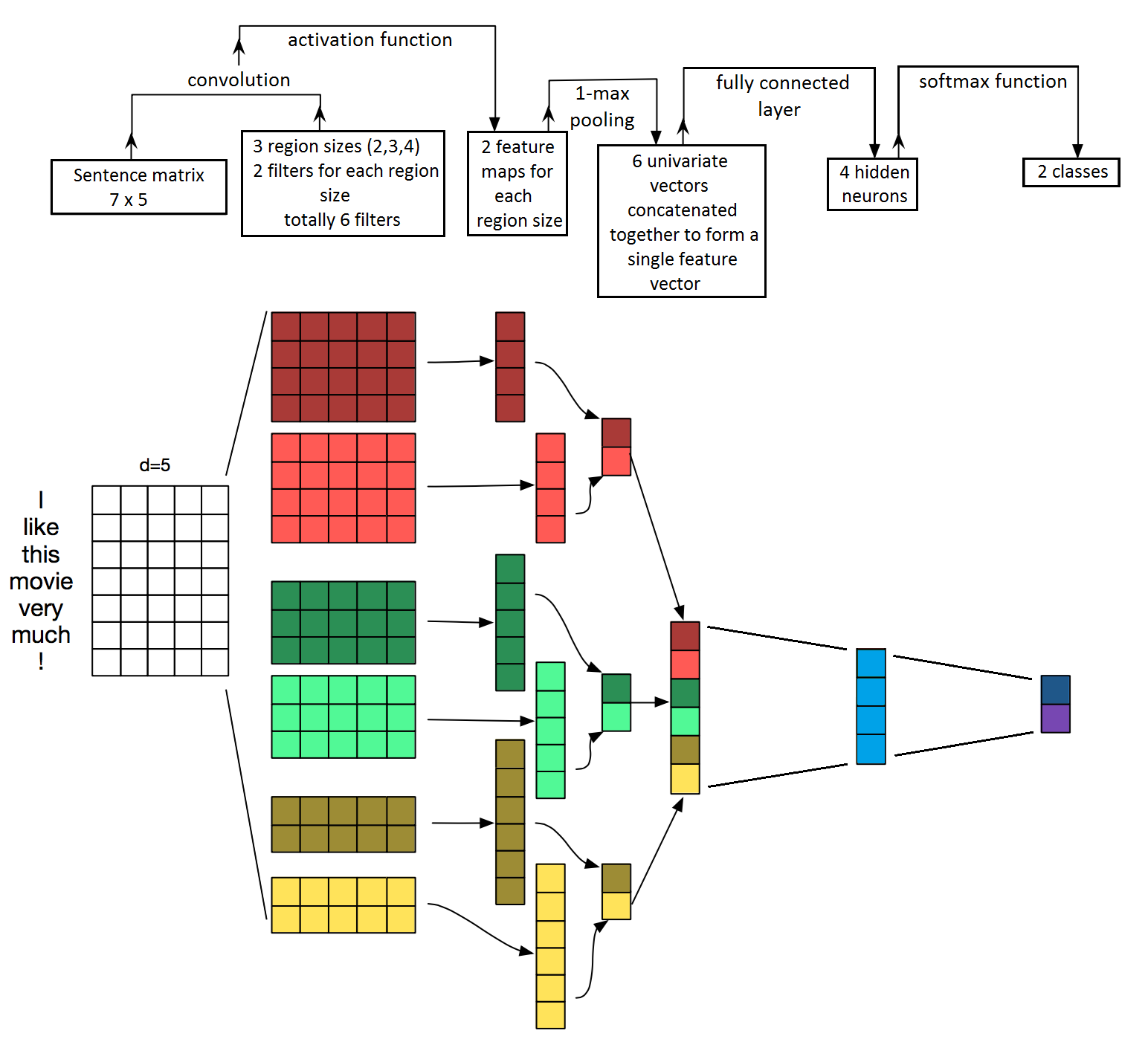}
\caption{Architecture of a smaller version of the CNN used.  Picture is taken from \protect\cite{zhang2015sensitivity} with minor modifications.}
\label{fig:conv}
\end{figure*}

\begin{figure*}[t]
\centering
\includegraphics[width=139.5mm]{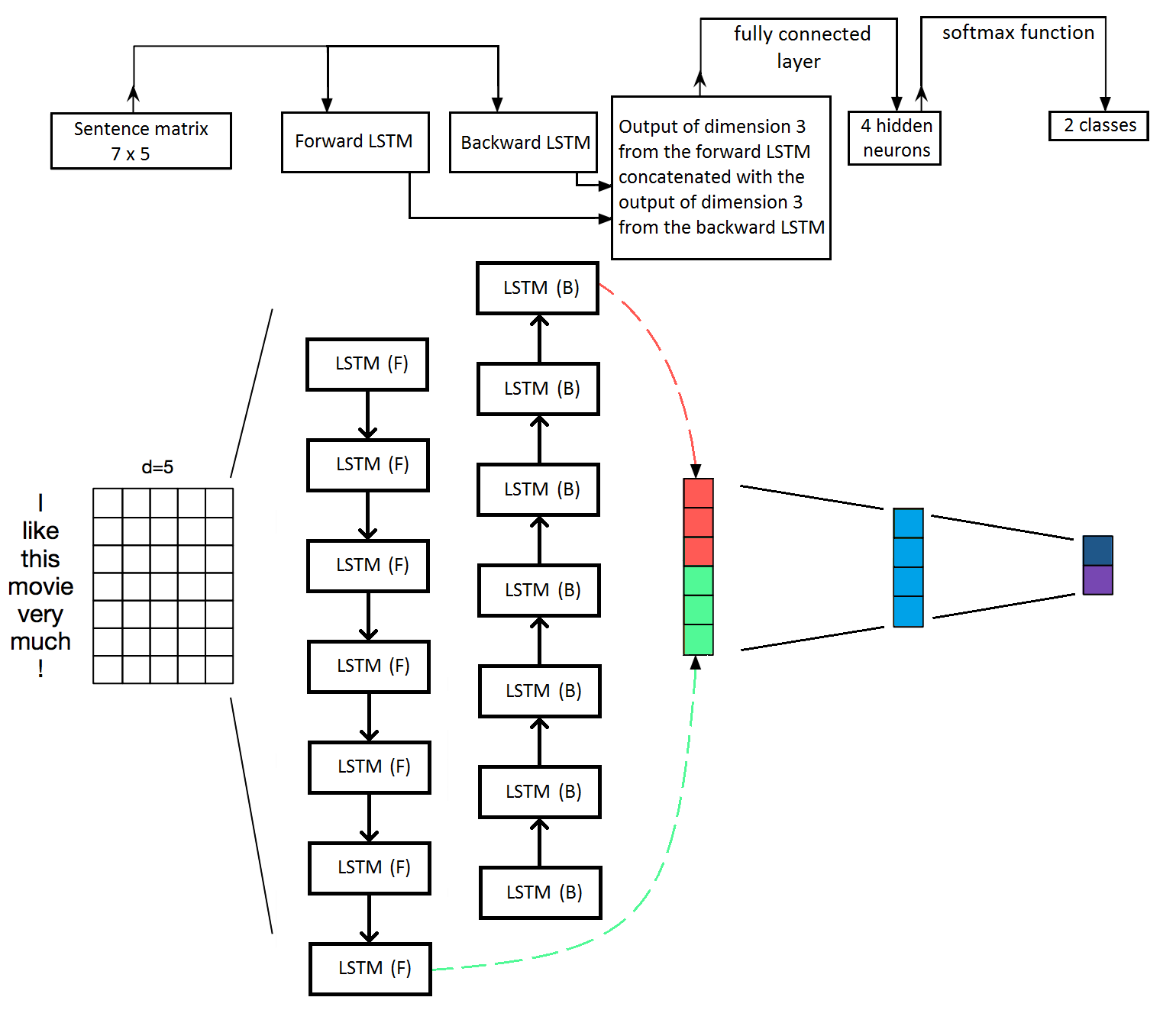}
\caption{Architecture of a smaller version of the bi-directional LSTM used.  Picture is inspired by Figure 1 of \protect\cite{zhang2015sensitivity}.}
\label{fig:lstm}
\end{figure*}

Let us now describe the architecture of the CNN we worked with.  Its architecture is almost identical to the CNN of \citet{kim2014convolutional}.  A smaller version of our model is illustrated on Fig. \ref{fig:conv}.   The input of the network are the tweets, which are tokenized into words.  Each word is mapped to a word vector representation, i.e. a word embedding, such that an entire tweet can be mapped to a matrix of size $s \times d$, where $s$ is the number of words in the tweet and $d$ is the dimension of the embedding space (we chose $d=200$).  We follow \citet{kim2014convolutional} zero-padding strategy such that all tweets have the same matrix dimension $X \in \mathbb{R}^{s'\times d}$, where we chose $s'=80$.  We then apply several convolution operations of various sizes to this matrix.  A single convolution involves a filtering matrix $w \in \mathbb{R}^{h\times d}$ where $h$ is the size of the convolution, meaning the number of words it spans.  The convolution operation is defined as 
\begin{eqnarray}
c_i &=& f \left( \sum_{j,k} w_{j,k} \left(X_{[i:i+h-1]} \right)_{j,k} + b \right)
\end{eqnarray}
where $b \in \mathbb{R}$ is a bias term and f(x) is a non-linear function, which we chose to be the relu function.  The output $c \in \mathbb{R}^{s'-h+1}$ is therefore a concatenation of the convolution operator over all possible window of words in the tweet.  Note that because of the zero-padding strategy we use, we are effectively applying wide convolutions \cite{kalchbrenner2014convolutional}. We can use multiple filtering matrices to learn different features, and additionally we can use multiple convolution sizes to focus on smaller or larger regions of the tweets.  In practice, we used three filter sizes (either $[1,2,3]$, $[3,4,5]$ or $[5,6,7]$ depending on the model) and we used a total of 200 filtering matrices for each filter size.\\

We then apply a max-pooling operation to each convolution $c_{\max} = \max(c)$.  The max-pooling operation extracts the most important feature for each convolution, independently of where in the tweet this feature is located. In other words, the CNN's structure effectively extracts the most important n-grams in the embedding space, which is why we believe these systems are good at sentence classification.  The max-pooling operation also allows us to combine all the $c_{\max}$ of each filter into one vector $\mathbf{c_{\max}} \in \mathbb{R}^{m}$ where m is the total number of filters (in our case $m=3\times 200 = 600$).  This vector then goes through a small fully connected hidden layer of size 30, which is then in turn passed through a softmax layer to give the final classification probabilities.  To reduce over-fitting, we add a dropout layer \cite{srivastava2014dropout} after the max-pooling layer and after the fully connected hidden layer, with a dropout probability of 50$\%$ during training.

\subsection{LSTM}

Let us now describe the architecture of the LSTM system we worked with.  A smaller version of our model is illustrated on Fig. \ref{fig:lstm}.  Its main building blocks are two LSTM units.  LSTMs are part of the recurrent neural networks (RNN) family, which are neural networks that are constructed to deal with sequential data by sharing their internal weights across the sequence.  For each element in the sequence, that is for each word in the tweet, the RNN uses the current word embedding and its previous hidden state to compute the next hidden state.  In its simplest version, the hidden state $h_t\in \mathbb{R}^{m}$ (where m is the dimension of the RNN, which we pick to be $m=200$) at time $t$ is computed by
\begin{eqnarray}
h_t &=& f\left( W_h\cdot x_t + U_h\cdot h_{t-1} + b_h\right)
\end{eqnarray}
where $x_t$ is the current word embedding, $W_h\in \mathbb{R}^{m\times d}$ and $U_h\in \mathbb{R}^{m\times m}$ are weight matrices, $b_h \in \mathbb{R}^{m}$ is a bias term and $f(x)$ is a non-linear function, usually chosen to be $\tanh$.  The initial hidden state is chosen to be a vector of zeros. Unfortunately this simple RNN suffers from the exploding and vanishing gradient problem during the backpropagation training stage \cite{hochreiter1998vanishing}.  LSTMs solve this problem by having a more complex internal structure which allows LSTMs to remember information for either long or short terms \cite{hochreiter1997long}.  The hidden state of an LSTM unit is computed by \cite{zaremba2014recurrent}
\begin{eqnarray}
f_t &=& \sigma\left(W_f\cdot x_t + U_f\cdot h_{t-1} + b_f\right) \nonumber \\
i_t &=& \sigma\left(W_i\cdot x_t + U_i\cdot h_{t-1} + b_i  \right) \nonumber\\
o_t &=& \sigma\left(W_o\cdot x_t + U_o\cdot h_{t-1} + b_o  \right) \nonumber\\
c_t &=& f_t \circ c_{t-1} \nonumber\\
& & + i_t\circ \tanh\left(W_c\cdot x_t + U_c\cdot h_{t-1} + b_c \right)\nonumber \\
h_t &=& o_t\circ \tanh\left(c_t \right)
\end{eqnarray}
where $i_t$ is called the input gate, $f_t$ is the forget gate, $c_t$ is the cell state,  $h_t$ is the regular hidden state,  $\sigma$ is the sigmoid function, and $\circ$ is the Hadamard product. \\

One drawback from the LSTM is that it does not sufficiently take into account post word information because the sentence is read only in one direction; forward.  To solve this problem, we use what is known as a bidirectional LSTM, which is two LSTMs whose outputs are stacked together.  One LSTM reads the sentence forward, and the other LSTM reads it backward. We concatenate the hidden states of each LSTM after they processed their respective final word.  This gives a vector of dimension $2m = 400$, which is fed to a fully connected hidden layer of size 30, and then passed through a softmax layer to give the final classification probabilities.  Here again we use dropout to reduce over-fitting; we add a dropout layer before and after the LSTMs, and after the fully connected hidden layer, with a dropout probability of 50$\%$ during training.

\section{Training}
\label{sec:train}

To train those models we had access to 49,693 human labeled tweets for subtask A, 30,849 tweets for subtasks (C, E) and 18,948 tweets for subtasks (B, D).  In addition to this human labeled data, we collected 100 million unique unlabeled English tweets using the Twitter streaming API.  From this unlabeled dataset, we extracted a distant dataset of 5 million positive tweets and 5 million negative tweets.  To extract this distant dataset we used the strategy of \citet{go2009twitter}, that is we simply associate positive tweets with the presence of positive emoticons (e.g. ``:)'') and vice versa for negative tweets.   Those three datasets (unlabeled, distant and labeled) were used separately in the three training stages which we now present. Note that our training strategy is very similar to the one used in \cite{severyn-moschitti:2015SemEval,deriu-EtAl:2016SemEval}.

\subsection{Pre-processing}

Before feeding the tweets to any training stage, they are pre-processed using the following procedure:
\begin{itemize}
\item URLs are replaced by the $<$url$>$ token.
\item Several emoticons are replaced by the tokens $<$smile$>$, $<$sadface$>$, $<$lolface$>$ or $<$neutralface$>$.
\item Any letter repeated more than 2 times in a row is replaced by 2 repetitions of that letter (for example, ``sooooo'' is replaced by ``soo'').
\item All tweets are lowercased.  
\end{itemize}

\subsection{Unsupervised training}

We start by using the 100 million unlabeled tweets to pre-train the word embeddings which will later be used in the CNN and LSTM.  To do so, we experimented with 3 unsupervised learning algorithms, Google's Word2vec \cite{mikolov2013efficient,mikolov2013distributed}, Facebook's FastText \cite{bojanowski2016enriching} and Stanford's GloVe \cite{pennington2014glove}.  Word2vec learns word vector representations by attempting to predict context words around an input word.  FastText is very similar to Word2vec but it also uses subword information in the prediction model.  GloVe on the other hand is a model based on global word-word co-occurrence statistics.  For all three algorithms we used the code provided by the authors with their default settings.  

\subsection{Distant training}

The embeddings learned in the unsupervised phase contain very little information about the sentiment polarity of the words since the context for a positive word (ex. ``good'') tends to be very similar to the context of a negative word (ex. ``bad'').  To add polarity information to the embeddings, we follow the unsupervised training by a fine tuning of the embeddings via a distant training phase.  To do so, we use the CNN described in sec. \ref{sec:desc} and initialize the embeddings with the ones learned in the unsupervised phase.  We then use the distant dataset to train the CNN to classify noisy positive tweets vs. noisy negative tweets.  The first epoch of the training is done with the embeddings frozen in order to minimize large changes in the embeddings.  We then unfreeze the embeddings and train for 6 more epochs.  After this training stage, words with very different sentiment polarity (ex. ``good'' vs. ``bad'') are far apart in the embedding space.  

\subsection{Supervised training}\label{subsec:sup}

The final training stage uses the human labeled data provided by SemEval-2017.  We initialize the embeddings in the CNN and LSTM models with the fine tuned embeddings of the distant training phase, and freeze them for the first $\sim 5$ epochs.  We then train for another $\sim 5$ epochs with unfrozen embeddings and a learning rate reduced by a factor of 10.  We pick the cross-entropy as the loss function, and we weight it by the inverse frequency of the true classes to counteract the imbalanced dataset. The loss is minimized using the Adam optimizer \cite{kingma2014adam} with initial learning rate of 0.001.  The models were implemented in TensorFlow and experiments were run on a GeForce GTX Titan X GPU.\\

To reduce variance and boost accuracy, we ensemble 10 CNNs and 10 LSTMs together through soft voting.  The models ensembled have different random weight initializations, different number of epochs (from 4 to 20 in total), different set of filter sizes (either $[1,2,3]$, $[3,4,5]$ or $[5,6,7]$) and different embedding pre-training algorithms (either Word2vec or FastText). \\

\begin{table*}[t]
 \centering
\begin{tabular}{|c|c|c|c|c|}
  \hline
  System & 2013 & 2014 & 2015 & 2016 \\
  \hline \hline
  Logistic regression on 1-3 grams baseline & 0.627 & 0.629 & 0.586 & 0.558 \\
  \hline
  CNN (word2vec, convolution size=[3,4,5]) & 0.715 & 0.723 & \textbf{0.688} & 0.643 \\
  \hline
  CNN (fasttext, convolution size=[3,4,5]) & 0.720 & 0.733 & 0.665 & 0.640 \\
  \hline
  CNN (glove, convolution size=[3,4,5]) & 0.709 & 0.714 & 0.660 & 0.637 \\
  \hline
  CNN (word2vec, convolution size=[1,2,3]) & 0.712 & 0.735 & 0.673 & 0.642 \\
  \hline
  CNN (word2vec, convolution size=[5,6,7]) & 0.710 & 0.732 & 0.676 & 0.646 \\
  \hline
  CNN (word2vec, convolution size=[3,4,5], no class weights) & 0.682 & 0.679 & 0.659 & 0.640 \\
  \hline
  CNN (word2vec, convolution size=[3,4,5], no distant training) & 0.698 & 0.716 & 0.660 & 0.636 \\
  \hline
  CNN (word2vec, convolution size=[3,4,5], no fully connected layer) & 0.715 & 0.724 & 0.683 & 0.641 \\
  \hline
  LSTM (word2vec) & 0.720 & 0.733 & 0.677 & 0.636 \\
  \hline
  LSTM (fasttext) & 0.712 & 0.730 & 0.666 & 0.633 \\
  \hline
  LSTM (glove) & 0.710 & 0.730 & 0.658 & 0.630 \\
  \hline
  LSTM (word2vec, no class weights) & 0.689 & 0.661 & 0.652 & 0.643 \\
  \hline
  LSTM (word2vec, no distant training) & 0.698 & 0.719 & 0.647 & 0.629 \\
  \hline
  LSTM (word2vec, no fully connected layer) & 0.719 & 0.725 & 0.675 & 0.634 \\
  \hline
  \hline
  Ensemble model & 0.725 & \textbf{0.748} & 0.679 & \textbf{0.648} \\
  \hline
  \hline
  Previous best historical scores & \textbf{0.728} & 0.744 & 0.671 & 0.633 \\
  \hline
\end{tabular}
  \caption{Validation results on the historical test sets of subtask A.  Bold values represent the best score for a given test set. The 2013 test set contains 3,813 tweets, the 2014 test set contains 1,853 tweets, the 2015 test set contains 2,392 tweets and the 2016 test set contains 20,632 tweets.  Word2vec, fasttext and glove refer to the choice of algorithm in the unsupervised phase. No class weights means no weights were used in the cost function to counteract the imbalanced classes. No distant training means that we used the embeddings from the unsupervised phase without distant training.  No fully connected layer means we removed the fully connected hidden layer from the network. Ensemble model refers to the ensemble model described in Sec. \ref{subsec:sup}. The previous best historical scores were collected from \protect\cite{SemEval:2016task4}.  They do not come from a single system or from a single team; they are the best previous scores obtained for each test set over the years. }
  \label{tab:cv}
\end{table*}

\section{Subtask specific tricks}
\label{sec:trick}

The models described in sec. \ref{sec:desc} and the training method described in sec. \ref{sec:train} are used in the same way for all five subtasks, with a few special exceptions which we now address.  Clearly, the output dimension differs depending on the subtask, for subtask A the output dimension is 3, while for B and D it is 2 and for subtask C and E it is 5.  Furthermore, for quantification subtasks (D and E), we use the probability average approach of \citet{bella2010quantification} to convert the output probabilities into sentiment distributions.\\

Finally for subtasks that have a topic associated with the tweet (B, C, D and E), we add two special steps which we noticed improves the accuracy during the cross-validation phase.  First, if any of the words in the topic is not explicitly mentioned in the tweet, we add those missing words at the end of the tweet in the pre-processing phase.  Second, we concatenate to the regular word embeddings another embedding space of dimension 5 which has only 2 possible vectors.  One of these 2 vectors indicates that the current word is part of the topic, while the other vector indicates that the current word is not part of the topic. \\ 

%Finally, while all subtasks benefit greatly from a cost function weighted by the inverse frequency of the true labels, we found that for subtask E this required some additional tunning.  Indeed, while we observed during the cross-validation phase that weights do help to acheive the best results, it requires many more training epochs (20 in total) compared to when we do not use weights. 

%Finally, while most tasks seemed to benifit greatly from a cost function weighted by the inverse frequency of the true labels, we found 2 exceptions to this rule.  For task D we found that it was better to leave the cost function unweighted.  Moreover, for task E, we found that while weights did help to acheive the best results in cross-validation, it required many more training epochs (20 in total) than for other tasks.  

%On the other hand, for task B we found that it was better to give an even bigger weight to the negative class than what the inverse frequency would indicate.  Indeed, since the metric for this task is the macro-average recall, we found that it was possible to overweight the negative class such that the negative recall is as large as the positive recall, which thus boosts their macro-average.  

\begin{table*}[t]
 \centering
\begin{tabular}{|c||c|c|c|c|c|c|}
  \hline
  System$/$System  &System 1 & System 2 & System 3 & System 4 & System 5 & System 6 \\
  \hline \hline
  System 1 & 1.0 & 0.95 & 0.97 & 0.97 & 0.93 & 0.91 \\
  \hline
  System 2 &  0.95 & 1.0 & 0.95 & 0.95 & 0.91 & 0.92 \\
  \hline
  System 3 & 0.97 & 0.95 & 1.0 & 0.96 & 0.92 & 0.91 \\
  \hline
  System 4 &  0.97 & 0.95 & 0.96 & 1.0 & 0.92 & 0.91 \\
  \hline
  System 5 &   0.93 & 0.91 & 0.92 & 0.92 & 1.0 & 0.95 \\
  \hline
  System 6 &   0.91 & 0.92 & 0.91 & 0.91 & 0.95 & 1.0 \\
  \hline
\end{tabular}
  \caption{Correlation matrix for the most important models. System 1: CNN (word2vec, convolution size=[3,4,5]), System 2:  CNN (fasttext, convolution size=[3,4,5]), System 3: CNN (word2vec, convolution size=[1,2,3]), System 4: CNN (word2vec, convolution size=[5,6,7]), System 5: LSTM (word2vec), System 6: LSTM (fasttext).}
  \label{tab:corr}
\end{table*}
\begin{table*}[t]
 \centering
\begin{tabular}{|c||c|c|c|c|}
  \hline
  Subtask  & Metric & Rank & BB$\_$twtr submission & Next best submission \\
  \hline \hline
  A & Macroaveraged recall & $1/38$ & 0.681 & 0.681 \\
  \hline
  B & Macroaveraged recall & $1/23$ & 0.882 & 0.856 \\
  \hline
  C & Macroaveraged mean absolute error & $1/15$ & 0.481 & 0.555 \\
  \hline
  D & Kullback-Leibler divergence & $1/15$ & 0.036 & 0.048 \\
  \hline
  E & Earth mover's distance &  $1/12$ &0.245 & 0.269 \\
  \hline
\end{tabular}
  \caption{Results on the 2017 test set.  The 2017 test set contains 12,379 tweets.  For a description of the subtasks and metrics used, see \protect\cite{SemEval:2017task4}.  For subtask A and B, higher is better, while for subtask C, D and E, lower is better.}
  \label{tab:res2017}
\end{table*}

\section{Results}
\label{sec:res}

Let us now discuss the results obtained from this system.  In order to assess the performance of each model and their variations, we first show their scores on the historical Twitter test set of 2013, 2014, 2015 and 2016 without using any of those sets in the training dataset, just like it was required for the 2016 edition of this competition.  For brevity, we only focus on task A since it tends to be the most popular one.  Moreover, in order to be consistent with historical editions of this competition, we use the average $F_1$ score of the positive and negative class as the metric of interest.  This is different from the macro-average recall which is used in the 2017 edition, but this should not affect the conclusions of this analysis significantly since we found that the two metrics were highly correlated.  The results are summarized in Table \ref{tab:cv}.  This table is not meant to be an exhaustive list of all the experiments performed, but it does illustrate the relative performances of the most important variations on the models explored here.\\

We can see from Table \ref{tab:cv} that the GloVe unsupervised algorithm gives a  lower score than both FastText and Word2vec.  It is for this reason that we did not include the GloVe variation in the ensemble model.  We also note that the absence of class weights or the absence of a distant training stage lowers the scores significantly, which demonstrates that these are sound additions.  Except for these three variations, the other models have similar scores. However, the ensemble model effectively outperforms all the other individual models.  Indeed, while these individual models give similar scores, their outputs are sufficiently uncorrelated such that ensembling them gives the score a small boost.  To get a sense of how correlated with each other these models are, we can compute the Pearson correlation coefficient between the output probabilities of any pairs of models, see Table \ref{tab:corr}.  From this table we can see that the most uncorrelated models come from different supervised learning models (CNN vs. LSTM) and from different unsupervised learning algorithms (Word2vec vs. FastText). \\

For the predictions on the 2017 test set, the system is retrained on all available training data, which includes previous years testing data.  The results of our system on the 2017 test set are shown on Table \ref{tab:res2017}.  Our system achieved the best scores on all of the five English subtasks.  For subtask A, there is actually a tie between our submission and another team (DataStories), but note that with respect to the other metrics (accuracy and $F_1^{PN}$ score) our submission ranks higher.

\section{Conclusion}
\label{sec:conc}

In this paper we presented the system we used to compete in the SemEval-2017 Twitter sentiment analysis competition.  Our goal was to experiment with deep learning models along with modern training strategies in an effort to build the best possible sentiment classifier for tweets.  The final model we used was an ensemble of 10 CNNs and 10 LSTMs with different hyper-parameters and different pre-training strategies.  We participated in all of the English subtasks, and obtained first rank in all of them. \\

For future work, it would be interesting to explore systems that combine a CNN and an LSTM more organically than through an ensemble model, perhaps a model similar to the one of \citet{stojanovski-EtAl:2016:SemEval}.  It would also be interesting to analyze the dependence of the amount of unlabeled and distant data on the performance of the models.

\section*{Acknowledgments}

We thank Karl Stratos, Anju Kambadur, Liang Zhou, Alexander M. Rush, David Rosenberg and Biye Li for their help on this project.

\bibliography{bb_twtr}
\bibliographystyle{acl_natbib}

\end{document}